%% file: main.tex
\documentclass[conference]{IEEEtran}

\usepackage[utf8]{inputenc}
\usepackage[T1]{fontenc}
\usepackage[english]{babel}
\usepackage{subscript}
\usepackage[table]{xcolor}

\usepackage{booktabs} 
\usepackage{array}

\usepackage{graphicx, float, subfigure, blindtext}

\newcommand\IEEEhyperrefsetup{
bookmarks=true,bookmarksnumbered=true,%
colorlinks=true,linkcolor={black},citecolor={black},urlcolor={black}%
}

\usepackage[\IEEEhyperrefsetup, pdftex]{hyperref}

\usepackage{mathtools}

\usepackage[nolist,nohyperlinks]{acronym}
\usepackage{cleveref}
\graphicspath{{images/}}

\usepackage{titlesec}
\titlespacing*{\section}{0pt}{*1}{*1}
\titlespacing*{\subsection}{0pt}{*1}{*1}
\renewcommand{\thesubsubsection}{\arabic{subsubsection}}
\titleformat{\subsubsection}[runin]{\itshape}{\thesubsubsection)}{1em}{}[:]
\titlespacing*{\subsubsection}{\parindent}{0pt}{*1}

\author{\textbf{\href{https://www.jua.ai/}{Jua.ai | AI for weather-dependent energy trading}}\\
\IEEEauthorblockA{%
Roberto Molinaro\textsuperscript{*}, Niall Siegenheim\textsuperscript{*}, Niels Poulsen\textsuperscript{*}, Jordan Dane Daubinet\textsuperscript{*}, Henry Martin\textsuperscript{*}, Mark Frey\textsuperscript{*}, Kevin Thiart\textsuperscript{*},\\
Alexander Jakob Dautel,  Andreas Schlueter, Alex Grigoryev, Bogdan Danciu,\\ Nikoo Ekhtiari, Bas Steunebrink, 
 Leonie Wagner, \\
Marvin Vincent Gabler (all affiliated with Jua.ai\textsuperscript{1})\\
\vspace{0.5em}
\textsuperscript{*}\textit{These authors contributed equally to this work.}}%
}

\title{EPT-2 Technical Report}
\begin{document}
\maketitle
\pagestyle{plain}
\thispagestyle{plain} 
\footnotetext[1]{\url{https://www.jua.ai/}}
\input{sections/abstract}
\input{sections/introduction}
\input{sections/arch}

\input{sections/bench_details}

\input{sections/results}

\input{sections/discussion}
\input{sections/acknowledgement}
\bibliographystyle{IEEEtran}
\bibliography{IEEEabrv,refs}

\clearpage
\input{sections/appendix}
\end{document}

%% file: sections/abstract.tex
\begin{abstract}
We present \mbox{EPT-2}, the latest iteration in our Earth Physics Transformer (EPT) family of foundation AI models for Earth system forecasting. \mbox{EPT-2} delivers substantial improvements over its predecessor, \mbox{EPT-1.5}, and sets a new state of the art in predicting energy-relevant variables-including 10m and 100m wind speed, 2m temperature, and surface solar radiation-across the full 0–240h forecast horizon. It consistently outperforms leading AI weather models such as Microsoft Aurora, as well as the operational numerical forecast system IFS HRES from the European Centre for Medium-Range Weather Forecasts (ECMWF).

In parallel, we introduce a perturbation-based ensemble model of \mbox{EPT-2} for probabilistic forecasting, called \mbox{EPT-2e}. Remarkably, \mbox{EPT-2e} significantly surpasses the ECMWF ENS mean-long considered the gold standard for medium- to long-range forecasting-while operating at a fraction of the computational cost. \mbox{EPT} models, as well as third-party forecasts, are accessible via the  \href{https://app.jua.ai}{\texttt{app.jua.ai}} platform.
\end{abstract}

%% file: sections/introduction.tex
\section{Introduction}
\label{section:intro}
The foundation of numerical weather prediction was established by Vilhelm Bjerknes in the early 20th century. Bjerknes proposed that atmospheric processes could be represented by a set of mathematical equations. Specifically, the fundamental laws of physics governing fluid dynamics and thermodynamics. By numerically solving these equations, it became possible to predict future states of the atmosphere based on current observational data.

Today, numerical weather prediction models simulate the atmosphere by dividing it into a three-dimensional grid and calculating changes in atmospheric variables like temperature, pressure, wind speed, and humidity at each grid point over time. These models require immense computational power and are continually refined to include more complex processes, such as cloud formation, radiation, and interactions between the atmosphere and the Earth's surface. Advanced data assimilation techniques incorporate real-time observational data from satellites, weather stations, and radar to enhance the accuracy of these models. Alongside, artificial intelligence and machine learning have been increasingly employed to optimize model parameters and improve predictive capabilities, marking a new era in weather forecasting.

Building upon these advancements, AI-based Earth system models have emerged that directly map raw observational data to predictions using end-to-end techniques. These models leverage deep learning to process vast datasets from satellites, radars, and sensors, capturing complex interactions among the atmosphere, oceans, land surfaces, and biosphere. By bypassing some traditional numerical methods, machine learning models aim to provide faster and more accurate forecasts. 

Examples of such AI-driven systems include \textit{Aurora}  \cite{bodnar2024foundationmodelearth}, a 1.3B-parameter transformer model trained on diverse datasets including ERA5, GFS, and IFS-HR, which currently sets the state of the art in global forecasting; \textit{GraphCast} from Google \cite{lam2023graphcastlearningskillfulmediumrange}, which introduced a GNN-based approach to model weather patterns; and \textit{AIFS} \cite{lang2024aifsecmwfsdatadriven}, ECMWF’s AI model that improves GraphCast by incorporating a sliding-window transformer processor. Other notable models include \textit{Pangu-Weather} \cite{bi2022panguweather3dhighresolutionmodel}, which uses 3D vision transformers; \textit{WeatherMesh-3} \cite{du2025weathermesh3fastaccurateoperational}, which performs latent-space forecasting; and \textit{Aardvark weather} \cite{vaughan2024aardvarkweatherendtoenddatadriven}, which improves ensemble reliability through learned distributional corrections.

Among the early efforts to operationalize these AI advances is Jua’s system. On March 1, 2023, Jua launched \textit{Vilhelm}~\cite{gabler2023vilhelmAIweather}, the first global AI weather model capable of producing native hourly forecasts. One month later, it introduced a 1\,km$\times$1\,km global precipitation product~\cite{gabler2023highresAIprecip}, followed by \mbox{EPT-1}, which extended the model’s capabilities to include high-resolution solar radiation forecasts. This was followed in October 2024 by \mbox{EPT-1.5} ~\cite{molinaro2024ept15technicalreport}, demonstrating superior performance to prior AI models such as \textit{GraphCast}, \textit{Pangu-Weather}, and \textit{FuXi-Weather}~\cite{sun2024fuxiweatherdatatoforecastmachine}.

Building on these developments, this report presents the latest iteration of the Jua models-\mbox{EPT-2} and its ensemble variant \mbox{EPT-2e}-and provides a comprehensive evaluation of their performance. 

The key contributions of this work are:

\begin{itemize}
    \item We introduce EPT-2, a compute-efficient, dynamically conditioned weather model that delivers state-of-the-art performance on deterministic forecasts.

    \item We present EPT-2e, an ensemble version of the model that considerably outperforms both Aurora and the ECMWF ENS across all key variables and lead times, achieving lower RMSE and CRPS throughout the full 0–240\,h forecast range.

    \item We highlight the operational advantages of the EPT-2 family, including 25\% faster inference, flexible lead time handling, and broad variable support.
\end{itemize}

%% file: sections/arch.tex
\section{EPT-2: Model Architecture and Operational Setting}
\label{section:arch}
The core of \mbox{EPT-2} is a predictive machine learning engine that maps sequences of past atmospheric states to future conditions. This approach mirrors the logic of traditional numerical weather prediction, but replaces hand-coded physical solvers with physical solvers learned from data.

This process can be more rigorously formalized by approximating the mapping:
\[
\Tilde{X}^{t+\Delta t} = f(X_t, X_{t-1}, \dots),
\]
which delineates the relationship between the current and previous states of the weather and the future state through a machine learning model. In this formulation, the model is denoted by $f_\theta$, where:
\[
X^{t+\Delta t} \approx \Tilde{X}^{t+\Delta t} = f_\theta(X_t, X_{t-1}, \dots).
\]
The model is trained by minimizing the following objective function:
\[
\mathcal{L}(X^t, X^{t+\Delta t}) = \frac{1}{C \cdot H \cdot W} \sum_{c=1}^C \sum_{i=1}^H \sum_{j=1}^W w_i \left| \tilde{X}_{c,i,j}^{t+\Delta t} - X_{c,i,j}^t \right| 
\]

Here, \(C\) denotes the number of weather variables, including temperature, wind velocity components, geopotential, etc., while \(H\) and \(W\) represent latitude and longitude, respectively. The weights \(w_i\) are given by:
\[
w_i = \frac{\cos(\text{lat}_i)}{\frac{1}{H} \sum \cos(\text{lat}_i)}
\]

This formula adjusts the importance of each term based on latitude, compensating for the varying density of grid points across latitudes in most geographic coordinate systems.

\subsection{Key Features}
The key features of \mbox{EPT-2} are outlined below: 

\begin{enumerate}
    \item \textbf{Foundational Earth Systems Model}: Like its predecessor \mbox{EPT-1.5}, \mbox{EPT-2} is an AI model trained on diverse datasets.
    
    \item \textbf{Any Lead Time Forecasting}: Unlike most current AI weather models that can only provide forecasts every six hours, \mbox{EPT-2} offers predictions for any  lead time $\Delta t$, and generates native hourly forecasts in an operational setting.

    \item \textbf{Probabilistic Forecasting}: \mbox{EPT-2} supports uncertainty-aware forecasting through an operational perturbation-based ensemble model, referred to as \mbox{EPT-2e}. This extends the core model’s capabilities and supports more robust decision-making under uncertainty.

\end{enumerate}

\subsection{Operational Specifications}

\begin{itemize}
    \item \textbf{Spatial Resolution}: \mbox{EPT-2} operates at a spatial resolution of 0.083 degrees or roughly 9 x 9 km at the equator, providing highly detailed forecasts. A high-resolution variant of EPT-2 has also been developed and may be discussed in future work; for comparability, we solely discuss the 9 x 9 km variant in this paper.
    \item \textbf{Temporal Resolution}: \mbox{EPT-2} currently runs at a temporal resolution of 1 hour up to 20 days into the future.
    \item \textbf{Forecast Frequency}: The model runs four times daily at 00:00, 06:00, 12:00, and 18:00 UTC. At each of these times, both an early version and a standard version are executed. The standard version becomes available with a typical 6-hour delay (like ECMWF IFS) but supports faster dissemination once initialized. The early version produces forecasts approximately 2.5 hours earlier\footnotemark. An enhanced high-refresh variant, designed for continuous, low-latency forecasting through direct assimilation of high-frequency observational data, is nearing public release and may be detailed in future work.
\end{itemize}

\footnotetext{\url{https://docs.jua.ai/models-and-products/dissemination-times}}

%% file: sections/bench_details.tex
\section{Benchmarks Details}
\label{section:benchmarking}
In this section, we outline the evaluation process for the performance of the EPT and third-party models. The pipeline is based on the well-established WeatherBench \cite{wb}.

\subsection{Ground Truth Datasets}
A fundamental aspect of evaluating model performance is the definition of a ground truth dataset. In the following, we describe three datasets commonly accepted in the weather community.

\textbf{ERA5.}
ERA5 \cite{era5dataset} is a state-of-the-art global atmospheric reanalysis dataset produced by the European Centre for Medium-Range Weather Forecasts (ECMWF). It is a gridded dataset and provides hourly estimates of a wide array of atmospheric, land, and oceanic variables from 1940 to the present, with a horizontal resolution of approximately 31~x~31 km at the equator (0.25 degrees).

\textbf{IFS HRES IC.}
The ECMWF High-Resolution Forecast (HRES) initial conditions (ICs) are computed from the most recent observational data. The high-resolution initial conditions serve as the starting point for ECMWF's  weather forecasts, which operate at a horizontal resolution of about 9~x~9 km at the equator (0.1 degrees). Although these are not direct observations, ICs are widely used in the research community due to their complete spatial coverage, ease of alignment, and historical use in training or validation pipelines.

\begin{figure}
  \centering
  \includegraphics[width=\columnwidth]{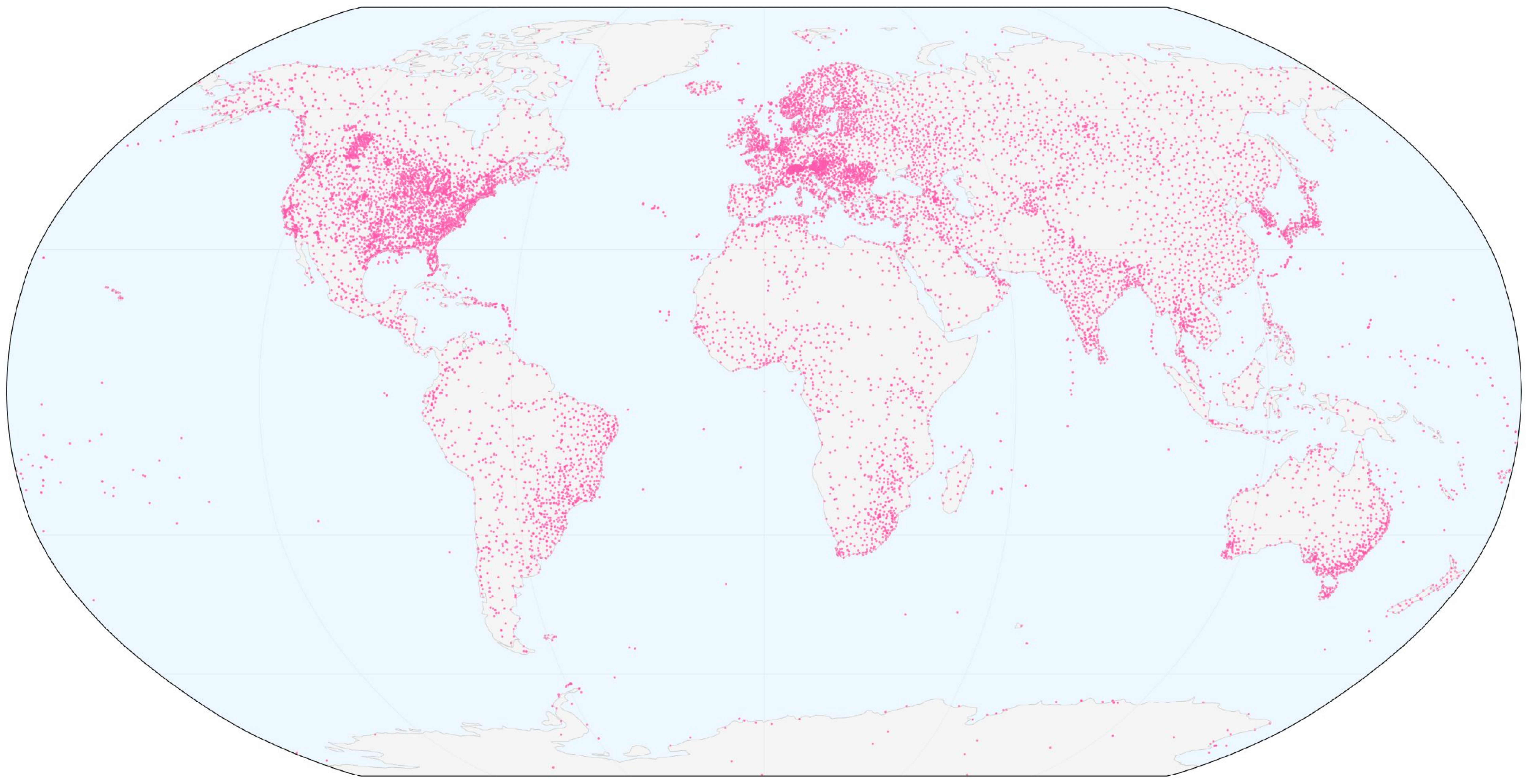}
  \caption{Spatial distribution of the weather stations for benchmarking}
  \label{fig:point_stations}
\end{figure}

\textbf{Weather Stations.} 
We use the WeatherReal‑ISD dataset~\cite{jin2024weatherrealbenchmarkbasedinsitu}, which aggregates global in-situ weather observations from over 14,000 stations as part of NOAA's Integrated Surface Database (ISD). These stations include traditional public networks such as SYNOP and METAR, reporting standard surface weather variables such as 2m air temperature, 10m wind speed, surface pressure, humidity, and precipitation. The data undergo extensive quality control and station unification procedures, ensuring consistency across both time and geography. Figure~\ref{fig:point_stations} shows the global distribution of stations used.  While spatially sparse and unevenly distributed, they offer a more neutral and fair ground truth for both AI and NWP models, free from the potential biases introduced when models are trained on datasets like ERA5. For this reason, station observations are widely regarded as a more faithful proxy for ground truth, and hence, serve as the primary evaluation target throughout this work. 
\textit{We emphasize that none of the models discussed in this work have been postprocessed or finetuned using weather station data, ensuring a fair comparison.}

\subsection{Models Evaluations}
Models are evaluated through two different types of benchmarkings:

\begin{enumerate}
    \item \textbf{Grid Benchmarks}: This approach compares the gridded model outputs with a gridded ground truth dataset, specifically IFS HRES IC.
    \item \textbf{Point Benchmarks}: This approach compares the model output at a specific point with an actual observation from WeatherReal  stations. The code used for running point benchmarks is available on GitHub at \href{https://github.com/juaAI/stationbench}{\texttt{github.com/juaAI/stationbench}}.
\end{enumerate}

Unless otherwise specified, all benchmark results are computed on a global scale.

\subsection{Methodology}
Next, we describe our benchmarking methodology more precisely.

\textbf{Initial Conditions.}
To ensure comparability, \mbox{EPT-2} is initialized using the same input data as the other models, providing a standardized baseline for evaluation.  All models are initialized at 00:00 and 12:00 UTC on a uniform schedule throughout 2023.

\textbf{Reference Forecast.}
As reference for deterministic evaluations, we choose the IFS HRES forecast produced by the European Centre for Medium-Range Weather Forecasts (ECMWF)~\cite{ecmwf}, benchmarked against its initial conditions and weather stations. IFS HRES is currently considered to be the world's most accurate numerical medium-range weather model. 

In the probabilistic evaluations, we also report the performance of ECMWF ENS mean, which averages forecasts across all members of the ECMWF ensemble prediction system (ENS). The ensemble mean offers a deterministic approximation that benefits from the spread of the ensemble, often improving skill over individual members, on the long-range time horizon.

\textbf{Evaluation Metrics.} 
We focus on two primary evaluation metrics: Root Mean Square Error (RMSE) and Continuous Ranked Probability Score (CRPS), which together assess both deterministic and probabilistic forecast quality.

\noindent\textit{Root Mean Square Error (RMSE).}
RMSE measures the average magnitude of forecast errors and is widely used in meteorology for evaluating deterministic predictions:
\begin{equation}
\text{RMSE} = \sqrt{\sum_{i=1}^{n} w_i \left( \tilde{X}_i - X_i \right)^2},
\end{equation}
where \( \tilde{X}_i \) is the predicted state (typically the deterministic forecast or the ensemble mean), \( X_i \) is the corresponding ground truth, and \( w_i \) are spatial weights (e.g., area-weighted). The spatial weighting $w_1$ is latitudes based for the grid benchmarks and $w_i=1$ for the point benchmarks. 

\noindent\textit{Continuous Ranked Probability Score (CRPS).}
CRPS is a strictly proper scoring rule for probabilistic forecasts, measuring the difference between the predicted cumulative distribution function (CDF) and the step function at the observation. For ensemble forecasts, the CRPS can be decomposed into two terms:
\begin{equation}
\text{CRPS} = \underbrace{\frac{1}{N} \sum_{i=1}^{N} w_i\left( \tilde{X}_i - X_i \right)^2}_{\text{Skill}} - \underbrace{\frac{1}{2N^2} \sum_{i=1}^{N} \sum_{j=1}^{N} w_i \left( \tilde{X}_i - \tilde{X}_j \right)^2}_{\text{Spread}},
\end{equation}
where \( \tilde{X}_1, \ldots, \tilde{X}_N \) are the predicted ensemble members. The first term measures accuracy (closeness to the observation), while the second captures ensemble spread (internal variability). 

A well-calibrated ensemble should balance spread and skill: if the ensemble is overconfident (too narrow), CRPS will penalize forecast misses; if it's too broad (underconfident), CRPS will penalize the lack of sharpness. Lower CRPS values indicate better overall probabilistic performance.

%% file: sections/results.tex
\section{Results}
\label{section:results}
In this section, we compare the performance of the \mbox{EPT-2} family against its predecessor \mbox{EPT-1.5}, as well as the current state-of-the-art weather model \textit{Aurora}~\cite{bodnar2024foundationmodelearth}, across two distinct evaluation settings:
\begin{itemize}

    \item \textbf{Deterministic Forecasting:} Evaluating forecast skill based on unperturbed ICs as input.
    \item \textbf{Probabilistic Forecasting:} Measuring the quality of uncertainty estimates and distributional outputs. Probabilistic forecasts for \mbox{EPT-2e} and Aurora are produced by introducing small perturbations to the ICs, with each perturbed state independently propagated through the model to generate the ensemble. 
\end{itemize}

\subsection{Deterministic Forecasting Results}
We evaluate the performance of \mbox{EPT-2} and baselines models using RMSE computed over the full 0--240\,h forecast range. Results are reported against both gridded (IC) and ground-based weather station observations. In addition to 10\,m wind speed, 100\,m wind speed, and 2\,m temperature, we also include results for surface solar radiation accumulated over six hours. This metric is only reported for \mbox{EPT-2} and \mbox{EPT-1.5}, as the publicly available Aurora model does not provide shortwave radiation output.

\begin{figure}[ht]
    \centering
    \includegraphics[width=1\linewidth]{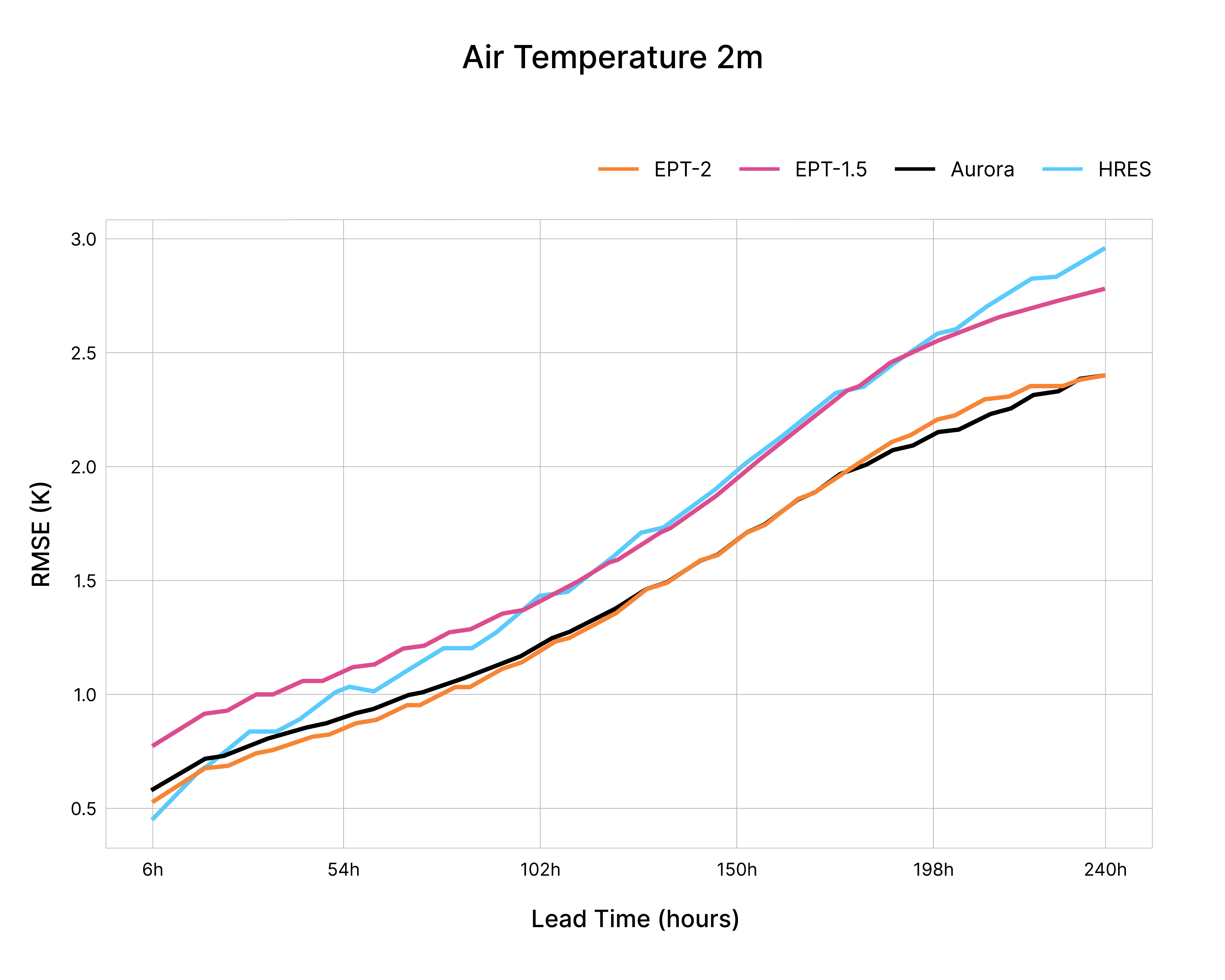}
    \caption{RMSE of 2m air temperature forecasts against ICs over the 0 – 240 h lead-time horizon.}
    \label{fig:2t_rmse_fientuned}
\end{figure}

\begin{figure}[ht]
    \centering
    \includegraphics[width=1\linewidth]{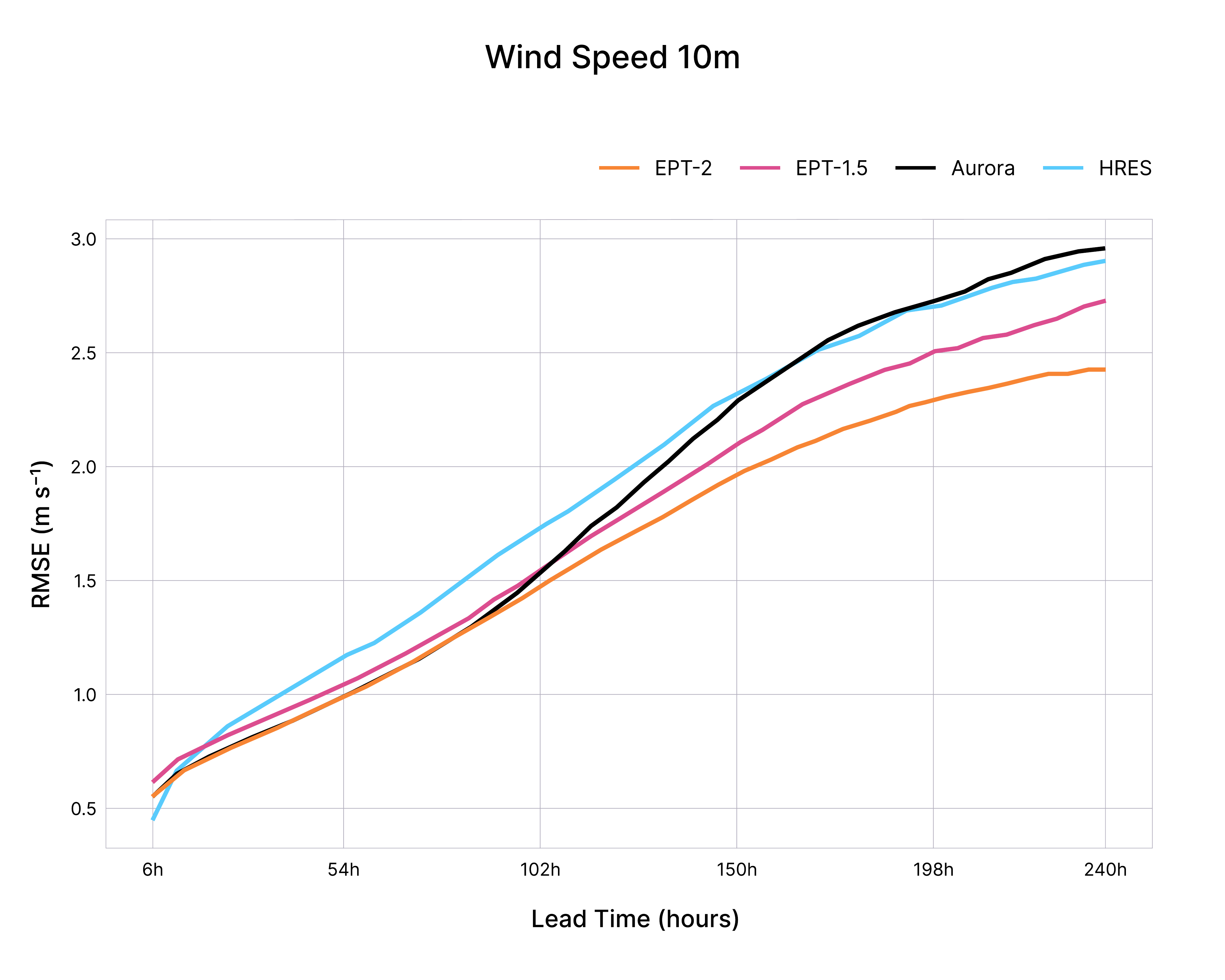}
    \caption{RMSE of 10m wind speed forecasts against ICs over the 0 – 240 h lead-time horizon.}
    \label{fig:10m_wind_finetuned}
\end{figure}

\begin{figure}[ht]
    \centering
    \includegraphics[width=\columnwidth]{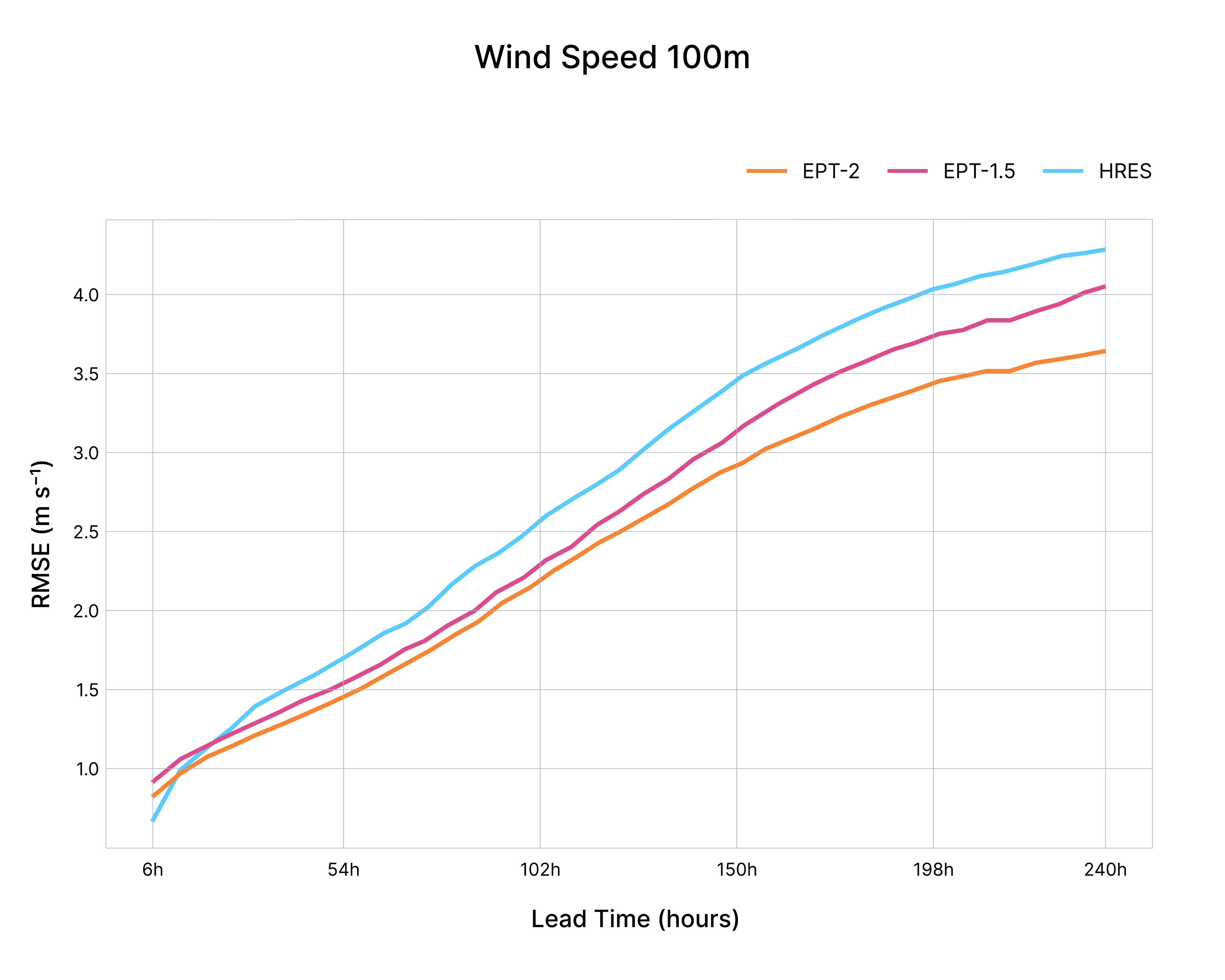}
    \caption{RMSE of 100m wind speed forecasts against ICs over the 0 – 240 h lead-time horizon.}
    \label{fig:100m_wind_finetuned}
\end{figure}

\begin{figure}[ht]
    \centering
    \includegraphics[width=1\linewidth]{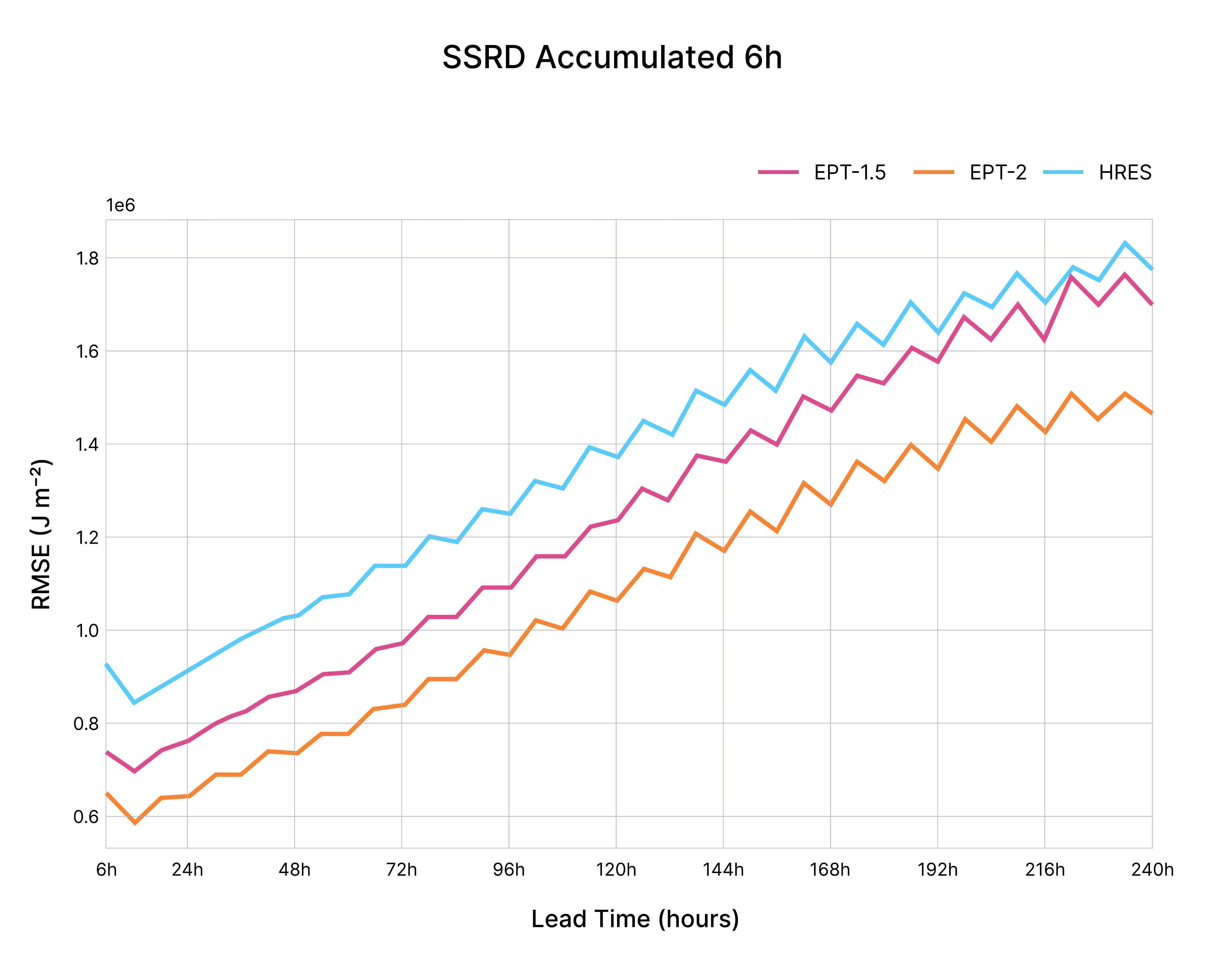}
    \caption{RMSE of solar radiation forecasts against ICs over the 0 – 240 h lead-time horizon.}
    \label{fig:solar_finetuned}
\end{figure}

\paragraph{\textbf{Evaluation against Initial Conditions.}} 
When evaluated against reference ICs (Figures \ref{fig:2t_rmse_fientuned}, \ref{fig:10m_wind_finetuned}, \ref{fig:100m_wind_finetuned}, \ref{fig:solar_finetuned}), \mbox{EPT-2} consistently outperforms \mbox{EPT-1.5} across all lead times and variables, including 10\,m wind speed, 100\,m wind speed, 2\,m air temperature, and surface solar radiation. For solar radiation, we use ERA5 as the reference, since SSRD is a diagnostic variable and not available in the HRES initial conditions.

Compared to Aurora, \mbox{EPT-2} achieves consistently lower RMSE on 10\,m wind speed across the entire 0--240\,h forecast horizon. For 2\,m temperature, \mbox{EPT-2} outperforms Aurora up to 130\,h, with a slight drop in relative performance at longer lead times. 

\mbox{EPT-2} also demonstrates substantial improvements over the HRES deterministic forecast beyond 12\,h across all variables considered. Notably, for solar radiation, \mbox{EPT-2} outperforms both \mbox{EPT-1.5} and HRES consistently throughout the full 0--240\,h range.

Additionally, Figures~\ref{fig:temp_pred} and~\ref{fig:wind_pred} show global snapshots of  2m temperature and 10m wind speed for \mbox{EPT-2}, Aurora and \mbox{EPT-1.5} at 1, 2, and 10 days lead time, alongside the initial condition. \mbox{EPT-2} delivers considerably sharper predictions at 10 days compared to Aurora, which produces visibly blurrier forecasts, and yet producing less accurate forecasts.

\begin{figure*}[t]
    \centering
    \begin{minipage}{0.48\linewidth}
        \includegraphics[width=\linewidth]{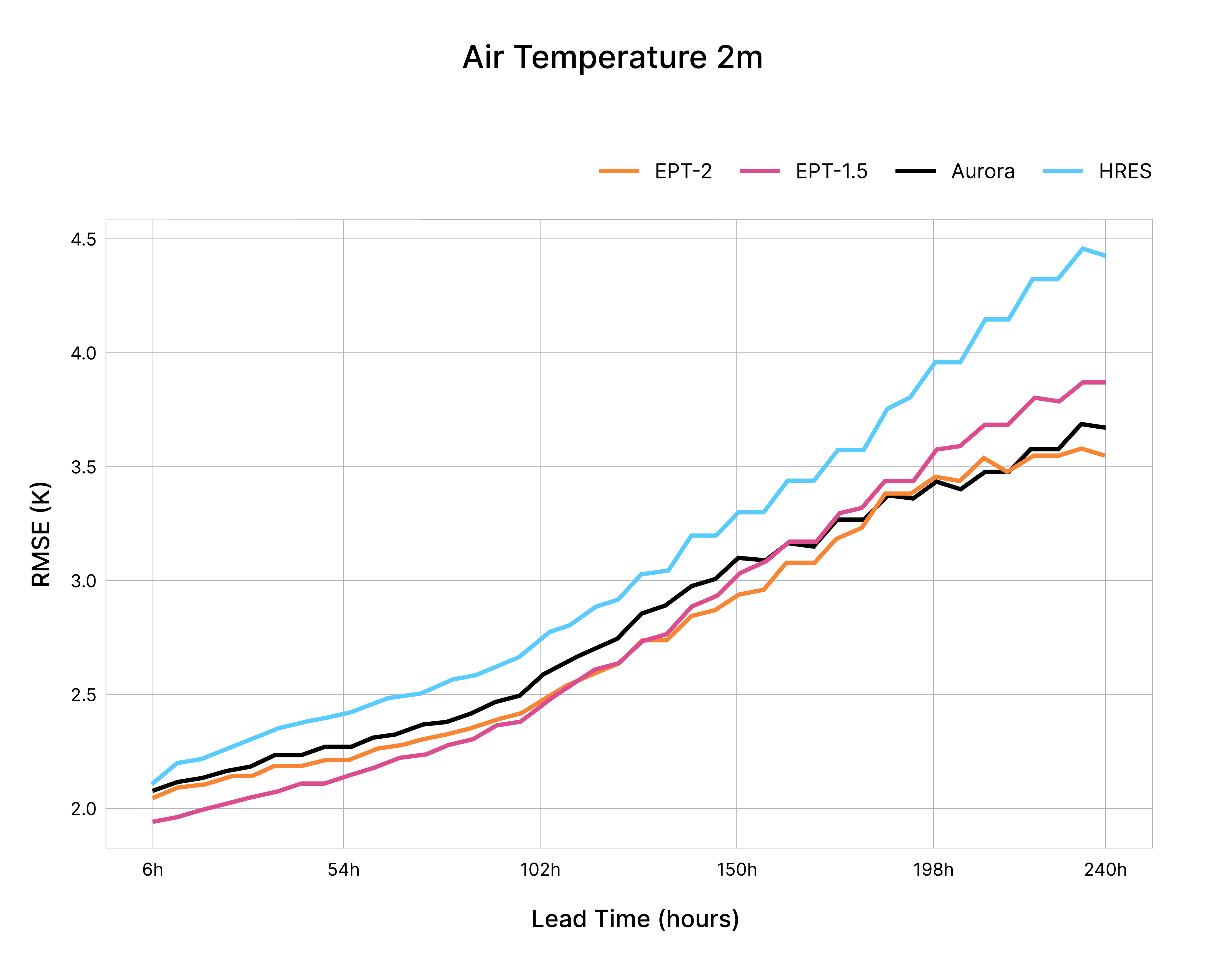}
        \caption{RMSE of 2m air temperature forecasts against in-situ weather station observations over the 0 – 240 h lead-time horizon.}
        \label{fig:2t_station}
    \end{minipage}
    \hfill
    \begin{minipage}{0.48\linewidth}
        \includegraphics[width=\linewidth]{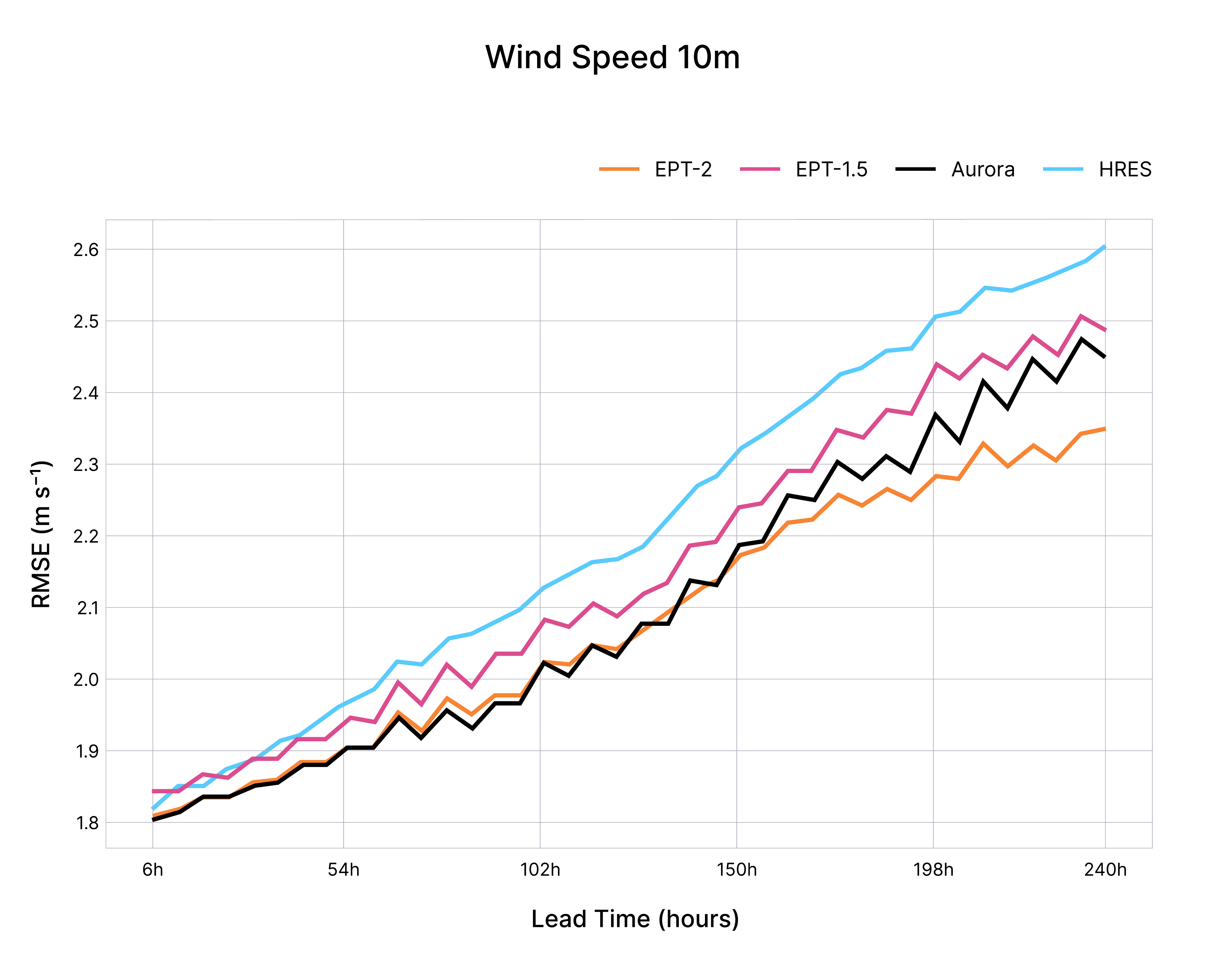}
        \caption{RMSE of 10m wind speed forecasts against in-situ weather station observations over the 0 – 240 h lead-time horizon.}
        \label{fig:10_wind_station}
    \end{minipage}
\end{figure*}

\paragraph{\textbf{Evaluation against Weather Stations.}}
Next, we evaluate model performance against in-situ weather station observations. \mbox{EPT-2} consistently outperforms \mbox{EPT-1.5} on 10m wind speed across all lead times (Figure~\ref{fig:10_wind_station}), confirming the improvements introduced in the latest version. For 2m air temperature (Figure~\ref{fig:2t_station}), \mbox{EPT-2} initially shows slightly higher RMSE than \mbox{EPT-1.5} up to 100 hours, but the difference narrows and reverses around 120 hours, after which \mbox{EPT-2} becomes more accurate.

Compared to \mbox{Aurora}, \mbox{EPT-2} performs similarly for short-term 10m wind forecasts, but achieves lower RMSE beyond 6 days. On temperature, \mbox{EPT-2} maintains an advantage over Aurora for almost the entire lead time range.

Across both variables and all lead times, all EPT variants outperform the HRES deterministic forecast, with the gap widening as lead time increases-highlighting the advantage of data-driven models in medium-range prediction.

Finally, to illustrate the EPT models’ capability for high-resolution temporal forecasting, Figures~\ref{fig:2t_ts} and~\ref{fig:10w_ts} show 4-day time series of temperature and wind speed at a Zurich location. These examples underscore the value of hourly predictions:
\begin{itemize}
\item Aurora misses most temperature peaks (Figure~\ref{fig:2t_ts}).
\item Around 36h, the EPT models capture a brief rise and fall in wind speed that Aurora overlooks-an edge that can be critical for energy applications (Figure~\ref{fig:10w_ts}).
\end{itemize}

\vspace{1em}

\begin{figure*}[t]
    \centering
    \begin{minipage}{0.48\linewidth}
        \includegraphics[width=\linewidth]{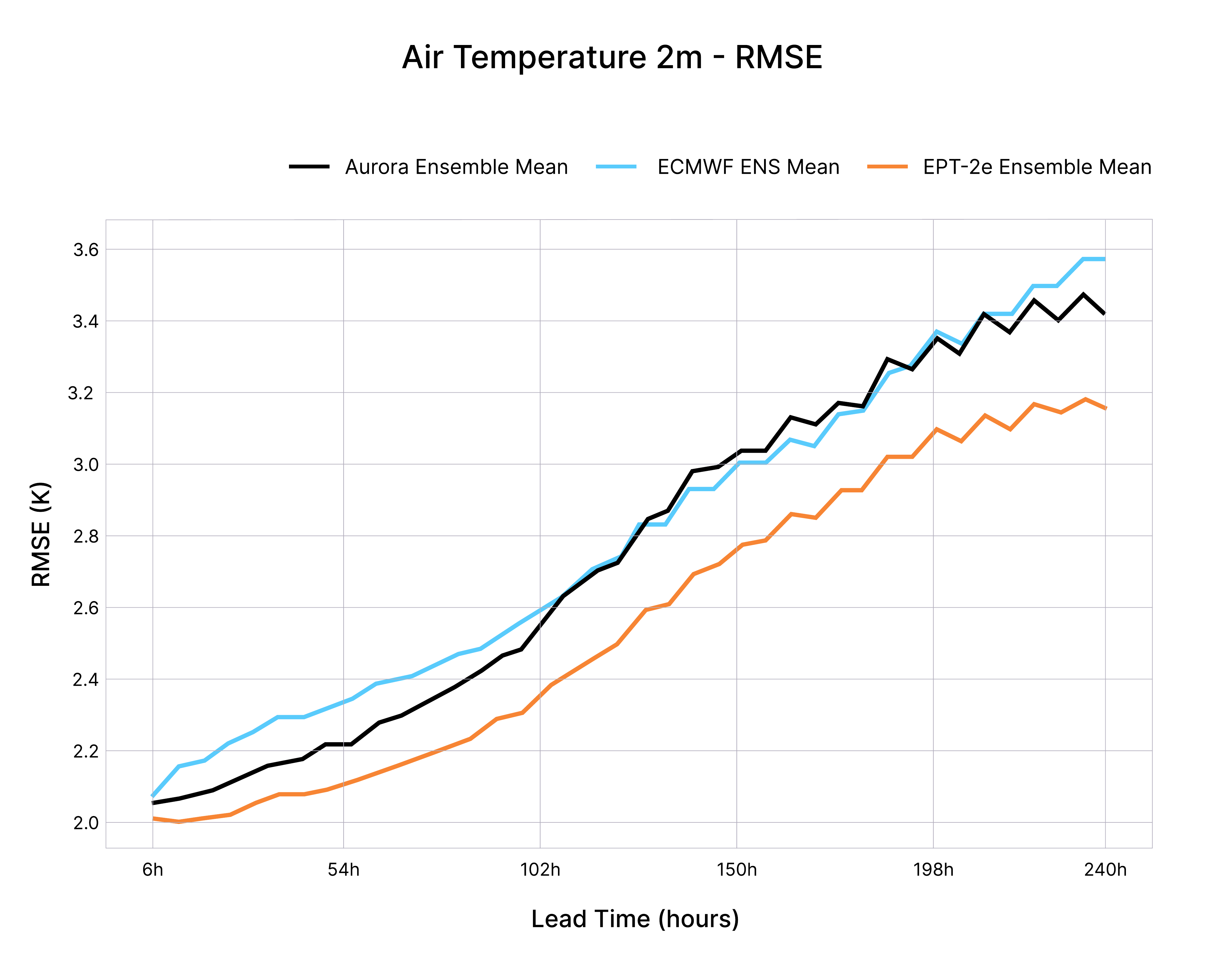}
        \caption{RMSE of 2m air temperature (ensemble mean) forecasts against in-situ weather station observations over the 0 – 240 h lead-time horizon.}
        \label{fig:2t_ens_rmse}
    \end{minipage}
    \hfill
    \begin{minipage}{0.48\linewidth}
        \includegraphics[width=\linewidth]{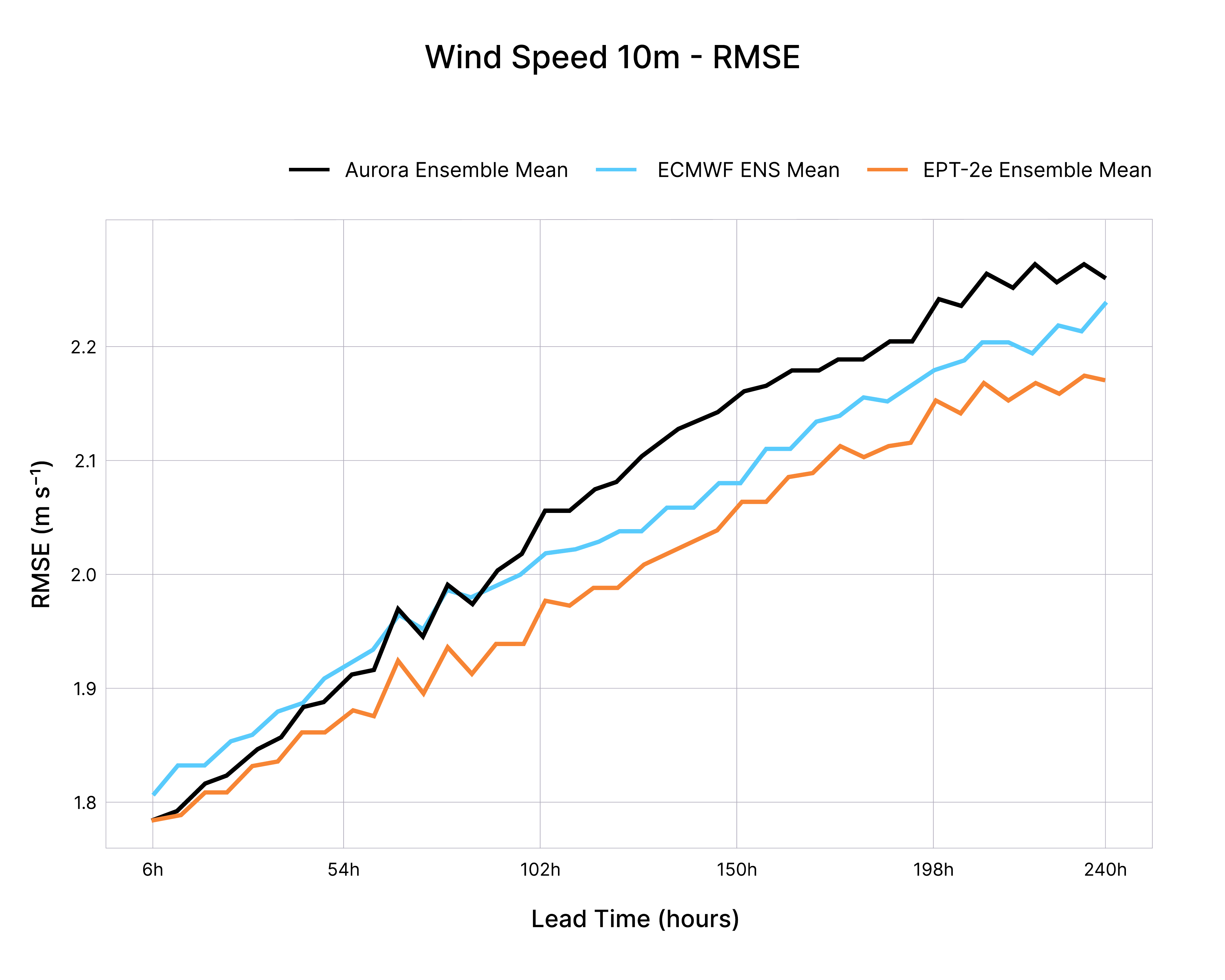}
        \caption{RMSE of 10m wind speed (ensemble mean) forecasts against in-situ weather station observations over the 0 – 240 h lead-time horizon.}
        \label{fig:10w_ens_rmse}
    \end{minipage}
\end{figure*}

\begin{figure*}[t]
    \centering
    \begin{minipage}{0.48\linewidth}
        \includegraphics[width=\linewidth]{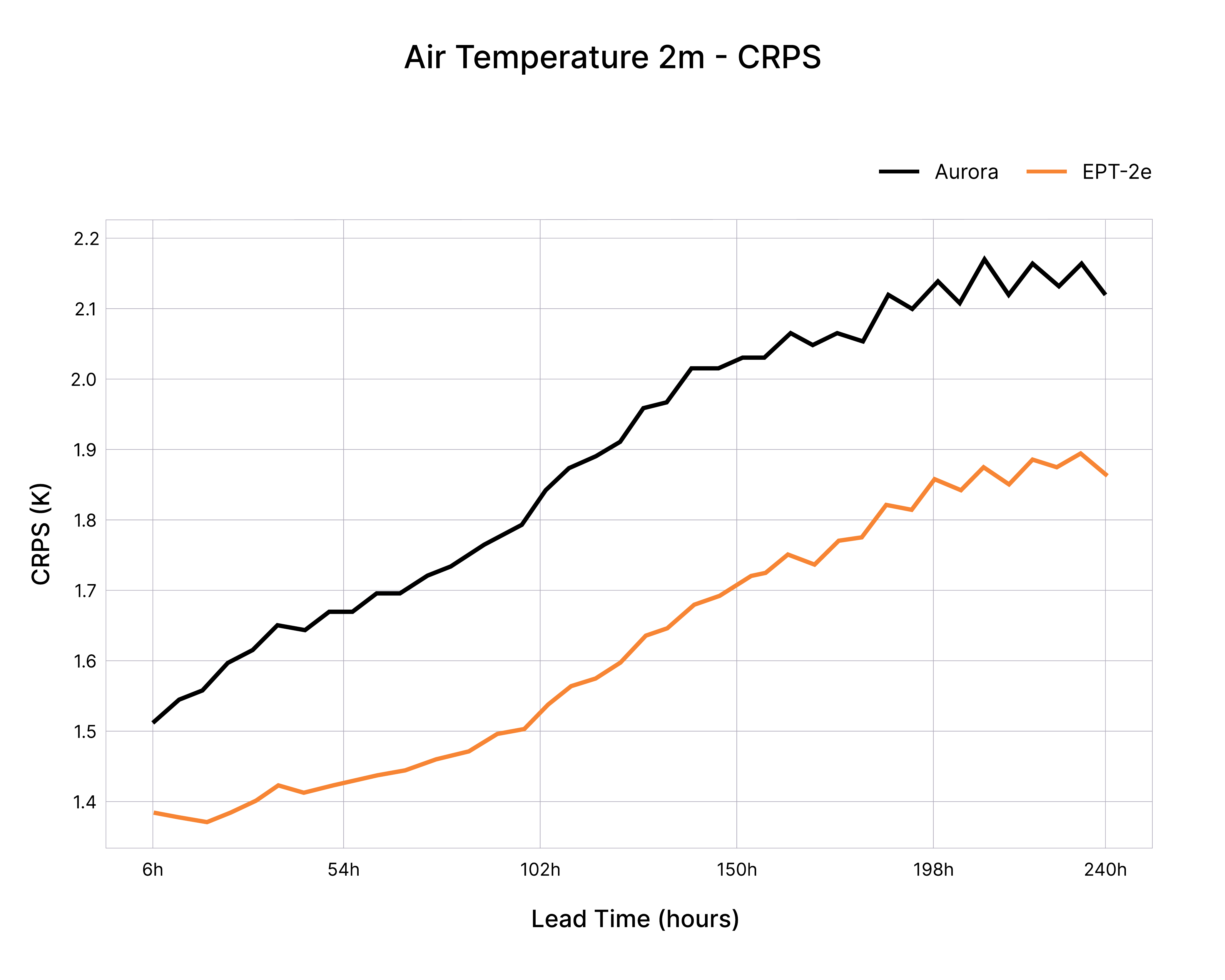}
        \caption{CRPS with weather station as reference for 2\,m air temperature (ensemble mean) across the 0--240\,h.}
        \label{fig:2t_ens_crps}
    \end{minipage}
    \hfill
    \begin{minipage}{0.48\linewidth}
        \includegraphics[width=\linewidth]{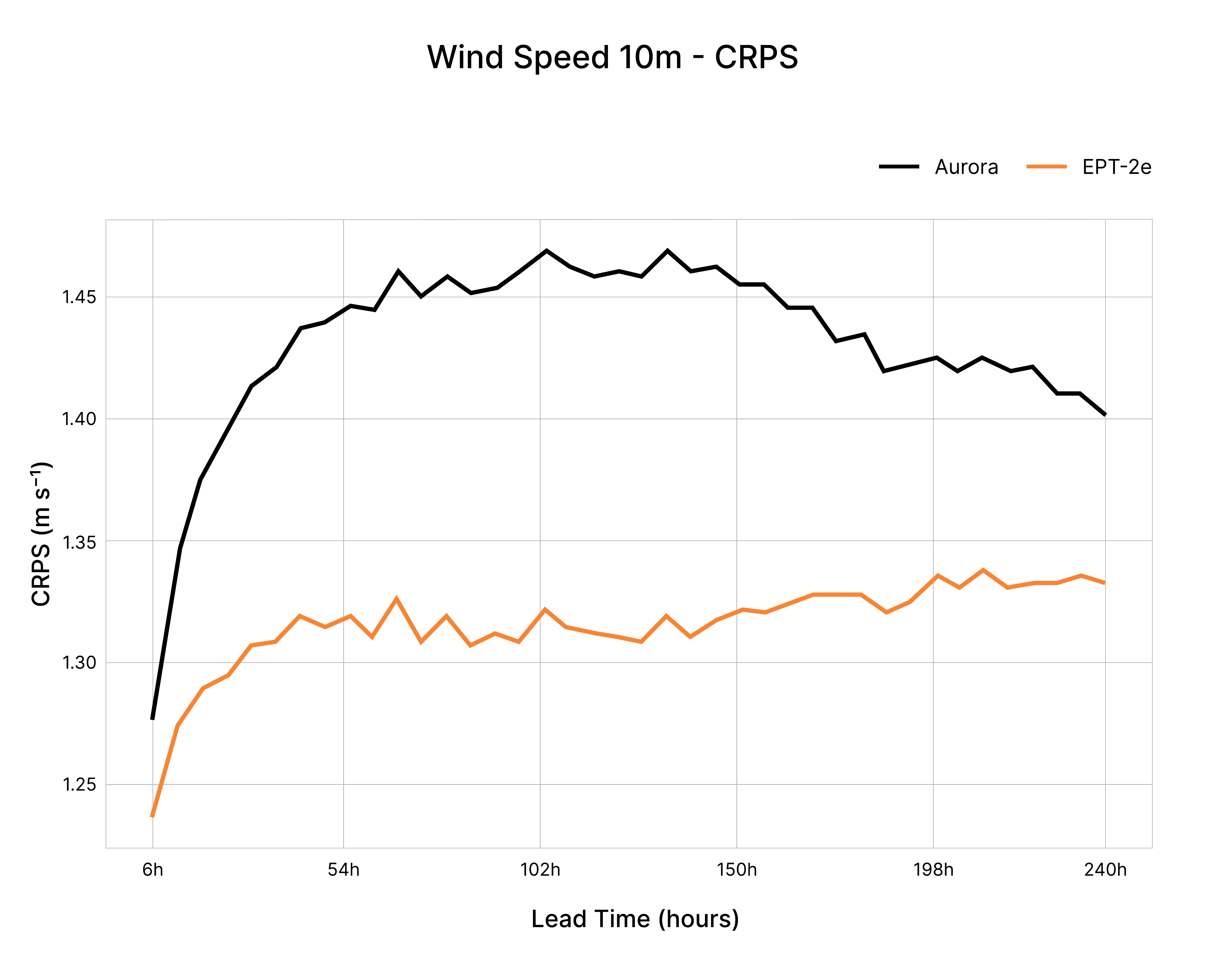}
        \caption{CRPS with weather station as reference for 10\,m wind speed (ensemble mean) across the 0--240\,h.}
        \label{fig:10w_ens_crps}
    \end{minipage}
\end{figure*}

\subsection{Probabilistic Forecasting Results.}
Next, we compare the ensemble performance of Aurora and \mbox{EPT-2e} using two complementary metrics: the root-mean-square error of the ensemble mean, and the continuous ranked probability score.
This evaluation is conducted exclusively against weather station observations, as they offer the most direct and fair measure of real-world forecast accuracy. Both AI models use ensembles composed of 10 members.

Figures~\ref{fig:2t_ens_rmse} and \ref{fig:10w_ens_rmse} present the RMSE of the ensemble mean for 2m air temperature and 10m wind speed across the 0–240h forecast horizon. \mbox{EPT-2e} clearly outperforms all other models, achieving the lowest RMSE at virtually every lead time.  Most notably, it consistently outperforms the ECMWF ENS mean-widely regarded as the gold standard for medium- to long-range forecasting-despite using only 10 members versus HRES’s 50. 

Figures~\ref{fig:2t_ens_crps} and \ref{fig:10w_ens_crps} confirm these findings from a probabilistic standpoint. \mbox{EPT-2e} achieves the lowest CRPS across almost the entire forecast range for both variables, with particularly large gains beyond 48h, demonstrating its ability to provide sharper and more reliable uncertainty estimates. 

This represents a significant step forward: \textit{a AI-based model not only matches, but in all the cases surpasses, the performance of a leading operational ensemble model-despite using far fewer members and requiring substantially lower computational cost.}

\subsection{Comparison with Aurora}

Finally, to provide a comprehensive view, we summarize key differences between \mbox{EPT-2} and Aurora across training cost, inference speed, and modeling flexibility:

\begin{itemize}
    \item \textbf{Forecast accuracy:} \mbox{EPT-2} consistently outperforms Aurora across variables and lead times.
    \item \textbf{Training efficiency:} \mbox{EPT-2} was pretrained on 8 H100 GPUs for 10 days, while Aurora required 32 A100 GPUs for 18 days. Despite using over 4$\times$ fewer GPUs and less total training time, \mbox{EPT-2} achieves comparable or superior forecast skill across most variables and lead times.
    To better understand scaling behavior and performance trade-offs, we also conducted extensive large-scale experiments, training model variants with up to 10 billion parameters. Nevertheless, the \mbox{EPT-2} version discussed in this paper remains dramatically more compute-efficient and cost-effective than other state-of-the-art models.
    
    \item \textbf{Inference speed:} \mbox{EPT-2} enables approximately 25\% faster inference than Aurora, making it better suited for time-critical operational deployments.

    \item \textbf{Lead time handling:} Aurora is trained to predict only fixed lead times (e.g., 6\,h), whereas \mbox{EPT-2} is trained with dynamic lead time conditioning. This allows \mbox{EPT-2} to generalize across arbitrary lead times without requiring multiple specialized output heads or model replications, reducing both training and inference complexity.
\end{itemize}

In summary, \mbox{EPT-2} matches or in the majority of the tasks outperforms Aurora in forecast skill while being significantly more efficient in terms of training compute, inference cost, and temporal flexibility.

%% file: sections/discussion.tex
\section{Discussion}
\label{section:disc}

\mbox{EPT-2} represents a substantial advancement in AI-based weather forecasting, delivering clear improvements over its predecessor \mbox{EPT-1.5} and other state-of-the-art AI models. In particular, it significantly enhances wind, temperature and solar radiation forecasting-addressing critical needs in energy production, including renewables, gas, and grid management. The model provides hourly global forecasts up to 20 days ahead, along with high-resolution ensemble prediction.

In operational benchmarks, \mbox{EPT-2} consistently outperforms ECMWF’s HRES deterministic model across multiple variables. Even more remarkably, its ensemble variant, \mbox{EPT-2e}, exceed the performance of the ECMWF ENS mean-long considered a benchmark for the entire time range requiring a fraction of the computational resources. These results are particularly impactful for applications that depend on accurate, high-frequency forecasts of temperature, solar radiation, and wind.

The combination of accuracy, temporal resolution, and efficiency positions \mbox{EPT-2} as a practical and powerful tool for real-time operational use, enabling more informed and responsive decision-making in weather-sensitive sectors.

%% file: sections/acknowledgement.tex
\section{Acknowledgements}
\label{section:ack}
We acknowledge the ECMWF for high-quality datasets and essential libraries like ecCodes. We thank the Max Planck Institute for Meteorology for their contributions to meteorological software tools that support our model development. We appreciate Stephan Rasp and the WeatherBench team for establishing a benchmark for evaluating AI weather models. We are grateful to the developers of the Zarr file format for enabling efficient storage and access to large datasets, and to Matthew Rocklin and the Dask development team for the Dask library, which has been instrumental in scaling our computational workflows. Finally, we thank our data partners and the broader meteorological community for their ongoing support in helping us evaluate and improve our models.

%% file: sections/appendix.tex
\onecolumn
\section{Supplementary Material}
\label{section:appendix}

\begin{figure*}[ht]
    \centering
    \includegraphics[width=0.8\linewidth]{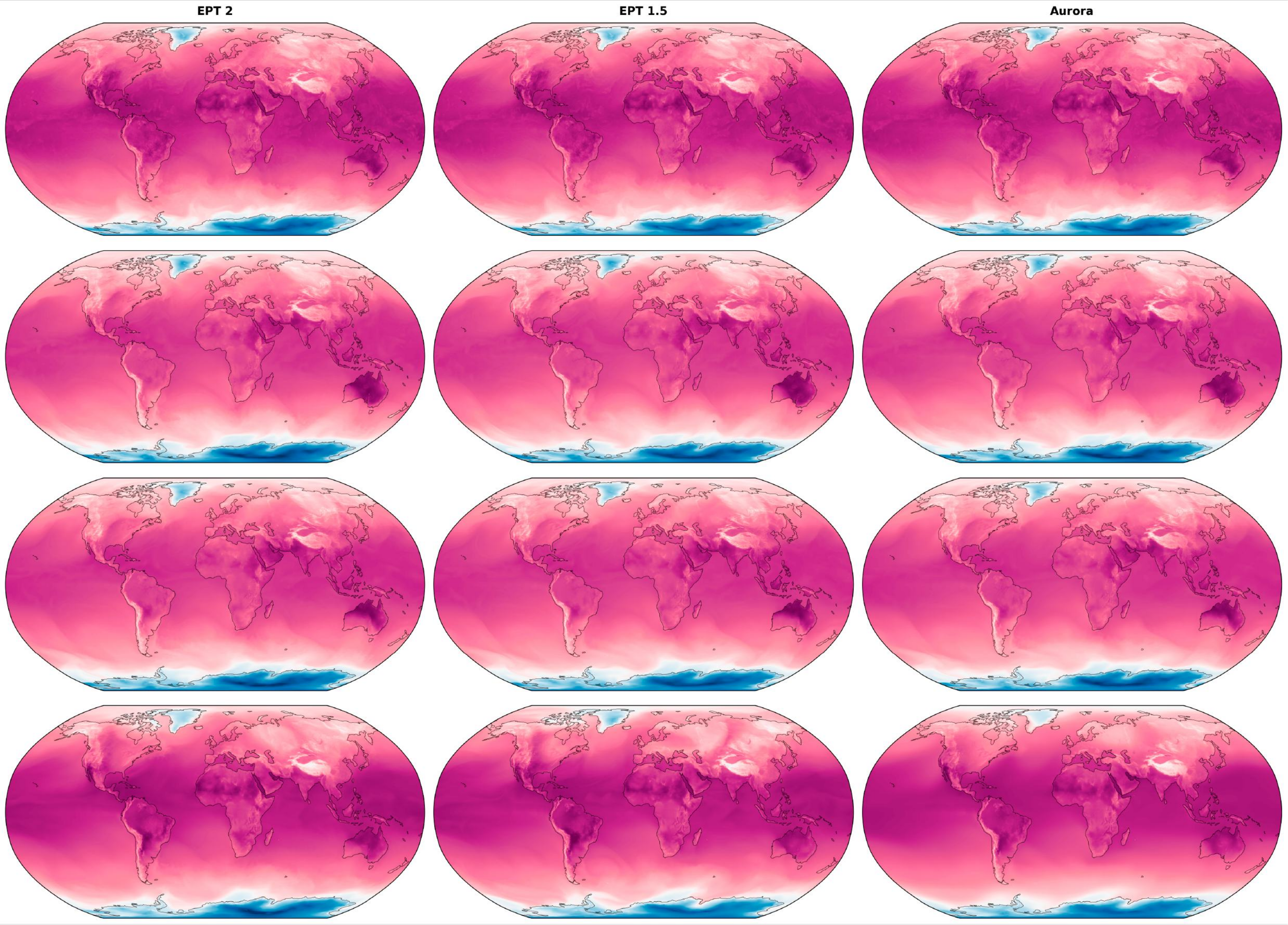}
    \caption{ 2m air temperature predictions from \mbox{EPT-2}, \mbox{EPT-1.5}, and Aurora on the 2023-10-01. The top row shows the initial condition (IC), while the second, third, and fourth rows display model predictions at 1-day, 2-day, and 10-day lead times, respectively.}
    \label{fig:temp_pred}
\end{figure*}

\begin{figure*}[ht]
    \centering
    \includegraphics[width=0.7\linewidth]{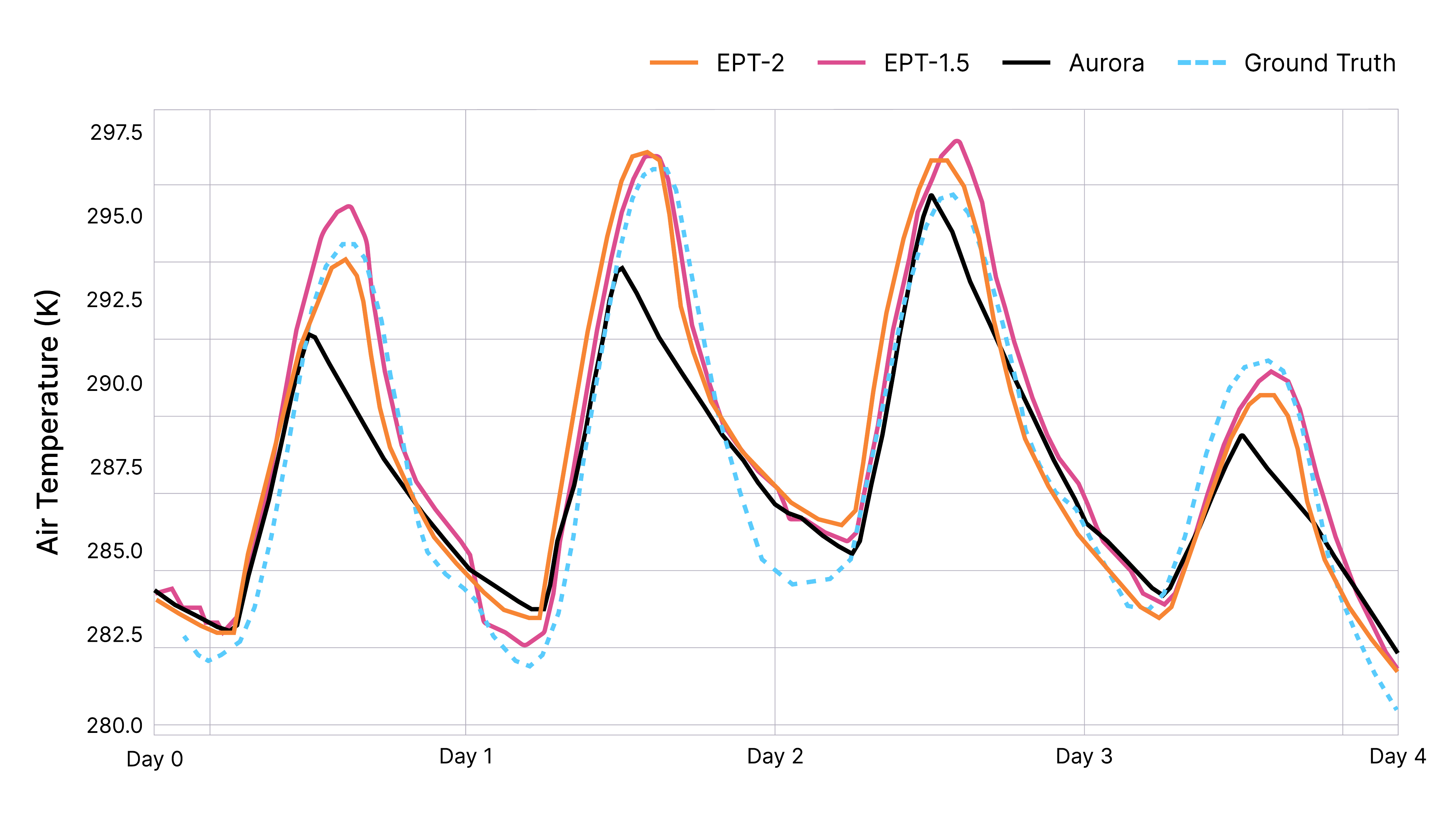}
    \caption{Time series of 2m temperature forecasts at a Zurich location on the 2023-10-01. EPT models provide hourly resolution, while Aurora outputs forecasts at six-hour intervals.}
    \label{fig:2t_ts}
\end{figure*}

\begin{figure*}[ht]
    \centering
    \includegraphics[width=0.8\linewidth]{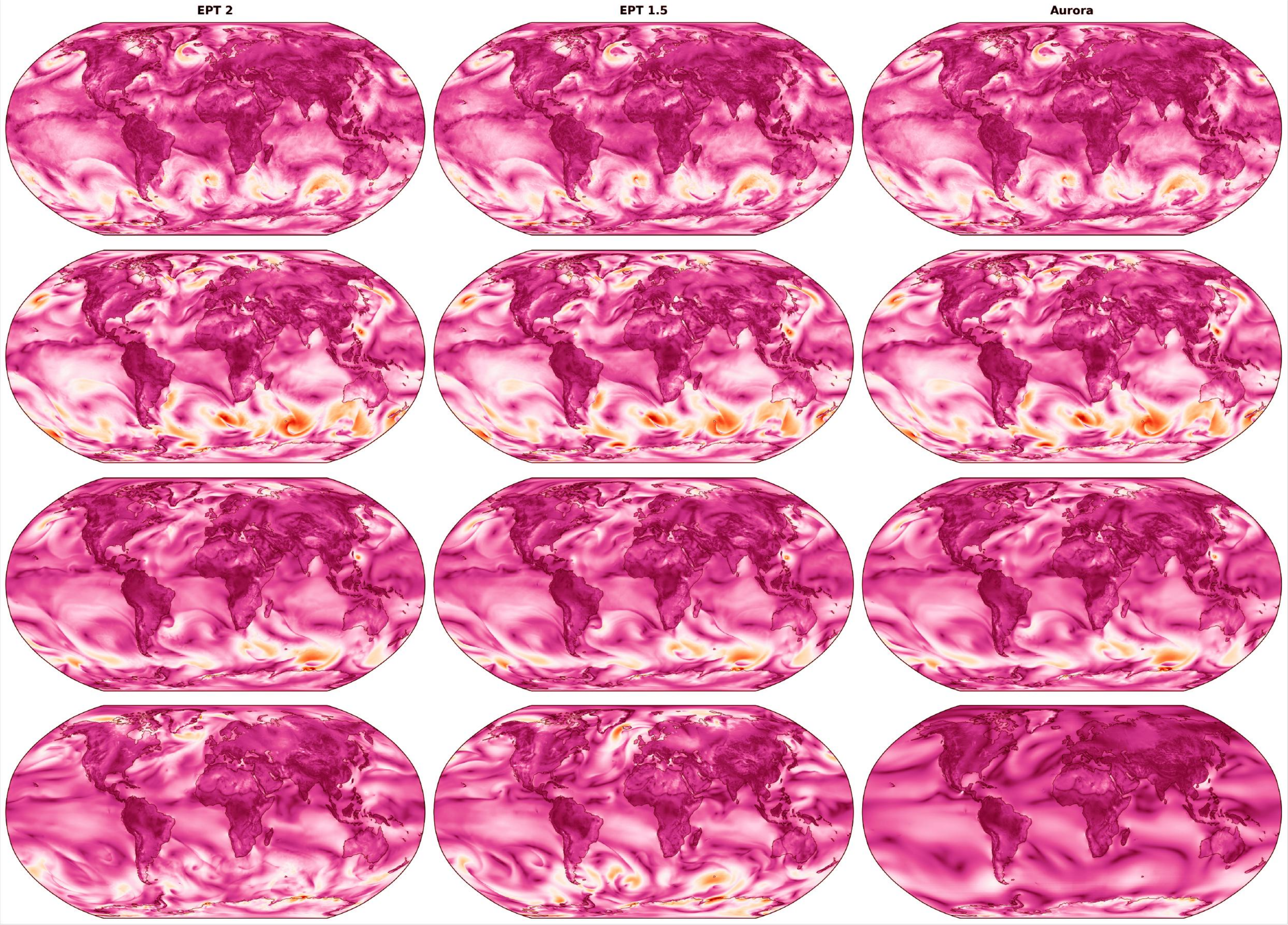}
    \caption{10m wind speed predictions from \mbox{EPT-2}, \mbox{EPT-1.5}, and Aurora on the 2023-10-01. The top row shows the initial condition (IC), while the second, third, and fourth rows display model predictions at 1-day, 2-day, and 10-day lead times, respectively.}
    \label{fig:wind_pred}
\end{figure*}

\begin{figure*}[ht]
    \centering
    \includegraphics[width=0.7\linewidth]{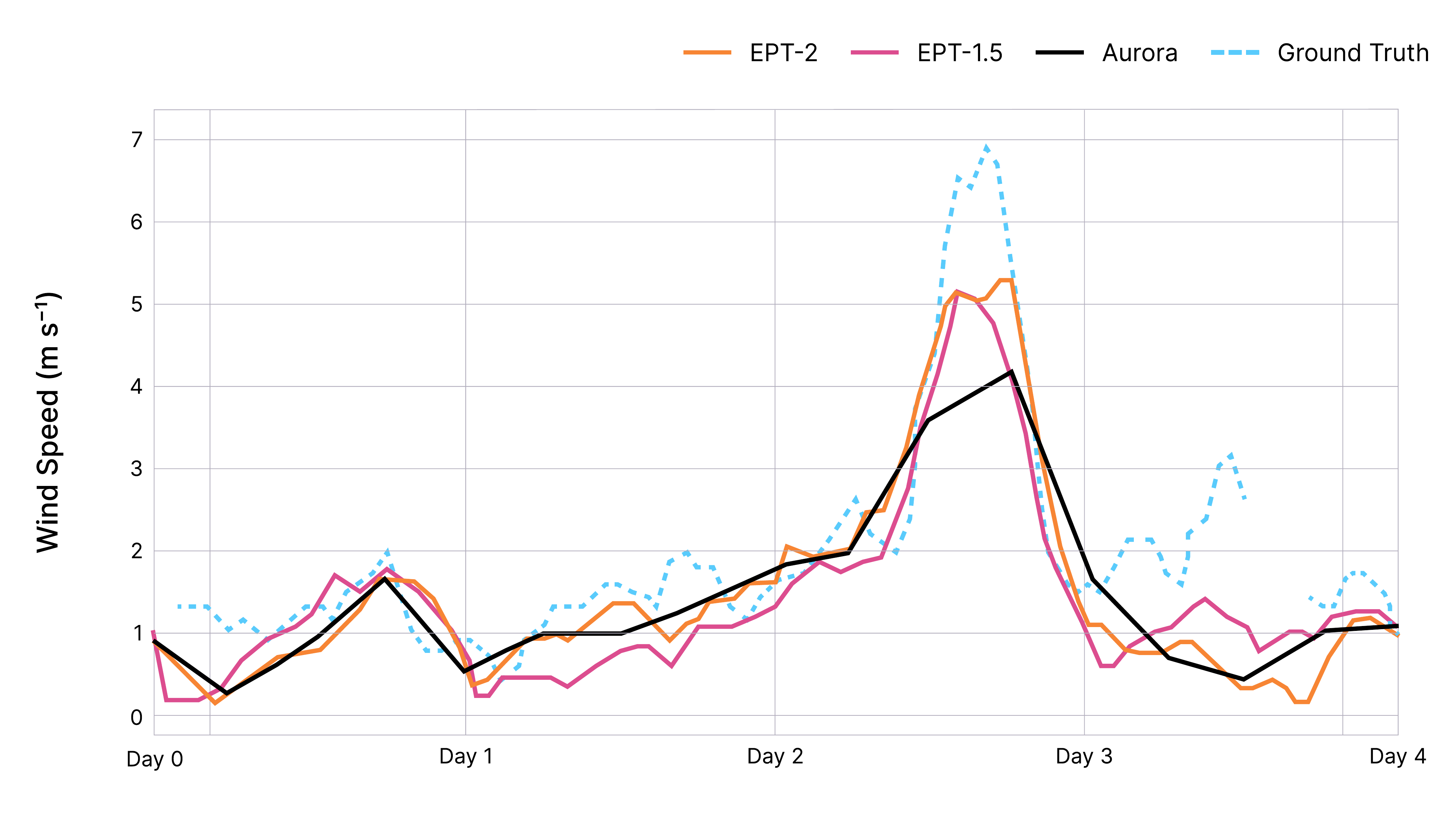}
    \caption{Time series of 10m wind speed forecasts at a Zurich location on the 2023-10-01. EPT models provide hourly resolution, while Aurora outputs forecasts at six-hour intervals.}
    \label{fig:10w_ts}
\end{figure*}